\documentclass[10pt,letterpaper]{article}
\usepackage{spconf}
\usepackage[hyphens]{url}
\usepackage{graphicx}
\usepackage{xcolor}
\usepackage[caption=false,font=small,labelformat=parens,labelsep=space]{subfig}
\usepackage{cite}
\usepackage{amsmath}
\usepackage{array}
\usepackage{booktabs}
\usepackage{placeins}
\usepackage[hidelinks]{hyperref}
\title{Protecting Private Values in Tool-Using LLM Agents via Recoverable Context Projection}

\name{Yu Li \qquad Qikun Cai \qquad Tao Huang \qquad Chen Hou}
\address{School of Computer and Big Data, Minjiang University \\
Fuzhou, China \\
liyu@stu.mju.edu.cn \qquad caiqikun@stu.mju.edu.cn \\
huang-tao@mju.edu.cn \qquad houchen@mju.edu.cn}

\begin{document}

\maketitle

\begin{abstract}
Memory-augmented and tool-using agents may expose exact private values when a remote LLM processes retrieved memory, tool actions, and observations. One-way masking limits initial exposure but removes values needed for trusted execution and may allow them to reappear in later observations. We propose Trustworthy Virtual Memory (TVM), a closed-loop runtime that exposes only projected context to the remote model, resolves exact values locally for trusted execution, and reprojects subsequent observations. We study whole-field and structure-preserving span-level projection, using a trusted local model to predict protected spans without access to the evaluation manifest. Experiments show that span-level projection preserves task-relevant structure, while recovery is necessary to turn protected reasoning into executable utility with low measured exposure.
\end{abstract}

\section{Introduction}
\label{sec:introduction}

Large language model (LLM) agents use retrieved demonstrations, historical actions, private records, and tool observations~\cite{lewis2020rag,yao2022react,shi2024ehragent}. When inference is remote, serialized requests may expose exact private values from retrieved memory or intermediate state; stored agent experience can itself be extracted~\cite{mextra}.

One-way masking redacts sensitive content before remote inference; PAPILLON evaluates such a baseline~\cite{siyan2025papillon}. This limits exposure in the initial request, but agent workflows span multiple requests. Tools may require exact identifiers removed by irreversible masking, and observations from a trusted executor may reintroduce those values in later requests. Protecting the initial prompt alone therefore does not cover the full execution cycle.

We introduce \emph{Trustworthy Virtual Memory} (TVM), a local runtime that separates remote-visible projected state from exact local execution state. TVM projects memory before remote inference, resolves protected references only before trusted execution, and reprojects observations before subsequent remote requests.

Projection must also preserve the structure needed for remote reasoning. Whole-field replacement can remove operators, relations, and action patterns along with protected values. We therefore study span-level projection, in which a trusted local model predicts protected spans and replaces only those spans with recoverable references.

We evaluate TVM on Memory-EHR and Memory-RAP using two remote providers.

Our contributions are threefold:
{\sloppy
\begin{itemize}
\item A closed-loop local--remote runtime that separates remote-visible context from exact local state through projection, validated recovery, and observation reprojection.
\item A recoverable projection interface instantiated at whole-field and span granularity; the span-level variant predicts protected spans locally without using the evaluation manifest.
\item Evidence that span-level projection preserves reasoning structure and that recovery is necessary for end-to-end executable utility.
\end{itemize}}

\section{Trustworthy Virtual Memory}
\label{sec:tvm}

We first define a detector-agnostic TVM runtime and then instantiate its projection interface at whole-field and span granularity.

\begin{figure*}[!t]
\centering
\includegraphics[width=\textwidth]{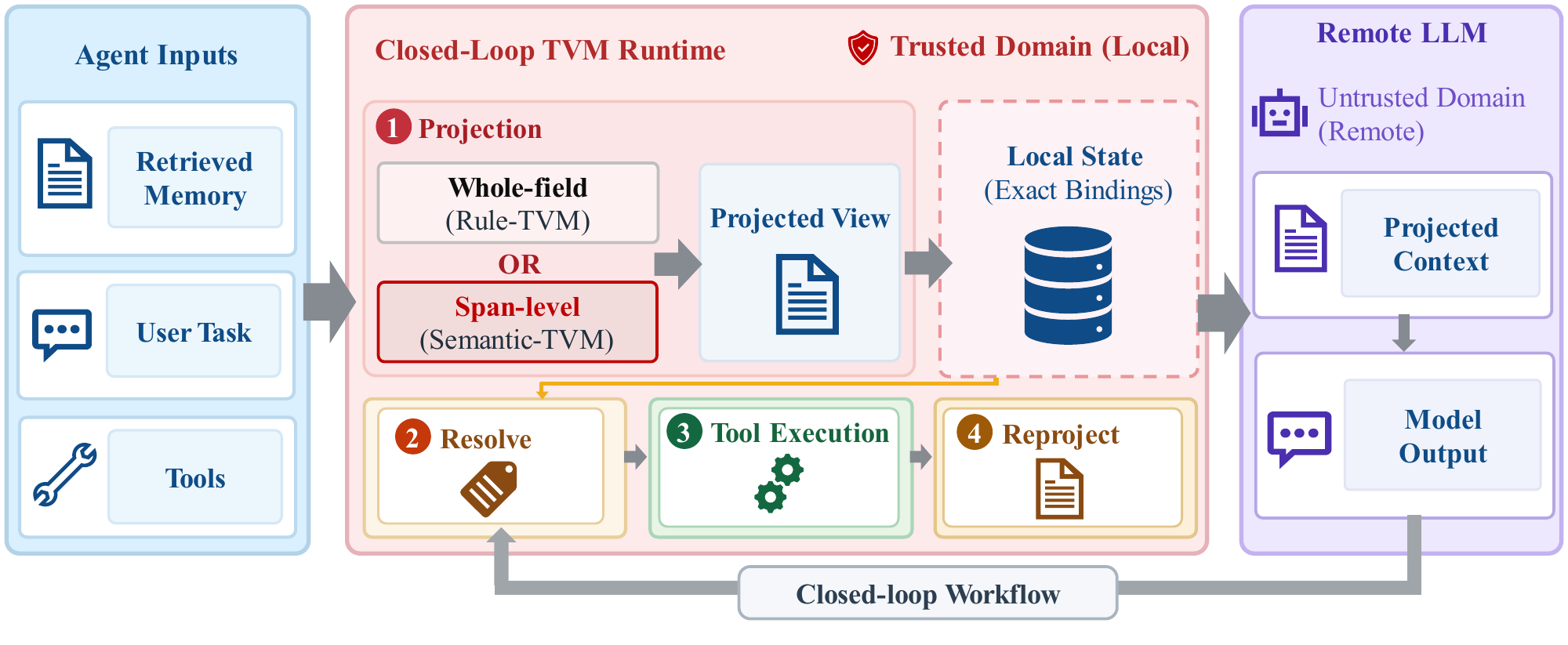}
\caption{TVM lifecycle with local recovery and observation reprojection.}
\label{fig:tvm-framework}
\end{figure*}

\subsection{Threat Model and Runtime}
\label{sec:abstraction}

\subsubsection*{Closed-Loop Runtime}
At round $t$, an agent with public task $x$ retrieves $R_t=\{m_{t,1},\ldots,m_{t,k}\}$ and queries an untrusted remote LLM $\mathcal{L}$. Before inference, TVM inserts a trusted local mediator $\mathcal{T}$,
\begin{equation}
  (\widetilde{R}_t,\rho_t)=\mathcal{T}_{\mathrm{project}}(R_t,x),
\end{equation}
where $\widetilde{R}_t$ is the remote-safe view and $\rho_t$ is local recovery state. For remote action $a_t$ and local executor $\mathcal{E}$,
\begin{equation}
  \begin{aligned}
    \widehat{a}_t &= \mathcal{T}_{\mathrm{resolve}}(a_t,\rho_{\leq t}),\quad
    o_t = \mathcal{E}(\widehat{a}_t),\\
    \widetilde{o}_t &= \mathcal{T}_{\mathrm{reproject}}(o_t,\rho_{\leq t}).
  \end{aligned}
\end{equation}
The remote model receives projected state but never recovery bindings; exact values are resolved only inside the trusted executor, and observations are reprojected before reuse. We capture every serialized outbound message for audit.

\subsubsection*{Trust Boundary}
We assume the remote provider can retain and inspect every serialized request, while retrieved memory and tool observations may contain protected values or adversarial instructions. The local mediator, recovery store, and executor are trusted and isolated from the provider. Our measured privacy objective is narrower: values designated protected for a trial must not occur in provider-bound messages.

\subsubsection*{Protected Information and Exposure Audit}
For trial $\tau$, $\mathcal{P}_{\tau}$ is an evaluation-only manifest of protected identifiers, record bindings, historical tool parameters, and their reappearances in tool results; it is never provided to the runtime projector. Runtime policy uses only scene, object type, and public-task context. Let $\mathcal{Q}_{\tau}$ denote the remote-bound messages and $\mathsf{Match}(v,q)$ detect an exact or conservatively normalized occurrence. A trial has no detectable exposure when
\begin{equation}
  \forall q \in \mathcal{Q}_{\tau},\;
  \forall v \in \mathcal{P}_{\tau}:
  \mathsf{Match}(v,q)=0,
\end{equation}

\subsubsection*{Recoverable Projection Interface}
Both projection variants implement the same reference contract. For an item $m$ and selected replacement units $S=\{s_j\}_{j=1}^{k}$,
\begin{equation}
  \widetilde{m}=m[s_j\leftarrow h_j]_{j=1}^{k},\qquad
  \rho(h_j)=(\ell_j,s_j,\mathrm{type}(s_j)),
\end{equation}
where $\ell_j$ records the structural path or character span of $s_j$. Whole-field projection selects complete field values, whereas span-level projection selects protected substrings within those fields. In both cases, $h_j$ serves as an opaque reasoning symbol for the remote model and an exact-value address for the local resolver; unreplaced content preserves compositional context.

Consider a retrieved EHR demonstration containing a lookup operation, its argument structure, and a protected patient identifier. Span-level projection replaces only the identifier with $h_7$, leaving the operation and surrounding constraints visible. The remote model can then emit \texttt{lookup(patient=}\ensuremath{h_7}\texttt{)}. Immediately before execution, TVM resolves $h_7$ through $\rho$ and executes the exact call locally. The resulting observation is reprojected before the next remote step. The reference thus connects protected remote reasoning to exact local execution throughout the tool trajectory.

\subsection{Rule-TVM: Whole-Field Projection}
\label{sec:rule-tvm}

Rule-TVM instantiates $\mathcal{T}_{\mathrm{project}}$ at whole-field granularity. A scene policy $\Pi_s$ selects protected structural paths; each value $v$ at path $p$ becomes a handle $h$, while
\[
  H_{\tau}[h]=(p,v,\mathrm{type}(v),\mathrm{policy}(p))
\]
remains in trial-local trusted state. Handles do not encode raw values. All variants share the validated resolver and reprojection path defined above.

\subsection{Semantic-TVM: Span-Level Projection}
\label{sec:semantic-tvm}

Whole-field projection may remove useful structure along with protected values. We therefore instantiate $\mathcal{T}_{\mathrm{project}}$ at span granularity as Semantic-TVM. For retrieved item $m_i$, task $x$, and frozen local scene adapter $A_s$, a trusted local model produces
\[
  (\widetilde{m}_i,\rho_i)=G_{\phi}(m_i,x,A_s),
\]
where $G_{\phi}$ predicts protected spans without the benchmark manifest. Each predicted span becomes a recoverable reference; $\rho_i$ retains its trial-scoped path, value, and type locally, while $\widetilde{m}_i$ retains the surrounding field or schema. The adapter conditions only $G_{\phi}$ within the trusted local runtime and is never serialized to the remote model; only $\widetilde{m}_i$ crosses the remote boundary. The adapter is induced once per run from clean probes containing neither protected values nor model outputs. 

As a granularity control, Classifier-TVM uses the same trusted local model to select protected fields but replaces each selected field in full.

\section{Experiments}
\label{sec:experiment}

\begin{table*}[!t]
\centering
\caption{Leak Rate $\downarrow$ (\% of fully captured trials) and Execution Success $\uparrow$ (\% of attempted trials) over three seeds; each cell is LR / ES. Bold marks the best Execution Success among defense methods on Memory-EHR.}
\label{tab:main-results}
\begin{tabular}{lcccc}
\toprule
\textbf{Task} & \multicolumn{2}{c}{\textbf{Memory-EHR}} & \multicolumn{2}{c}{\textbf{Memory-RAP}} \\
\cmidrule(lr){2-3}\cmidrule(lr){4-5}
\textbf{Method} & \textbf{DeepSeek} & \textbf{MiniMax} & \textbf{DeepSeek} & \textbf{MiniMax} \\
\midrule
No Defense & 78.50 / 97.83 & 78.50 / 94.50 & 100.00 / 100.00 & 100.00 / 99.56 \\
Masking & 0.34 / 59.33 & 0.00 / 44.83 & 6.00 / 100.00 & 22.67 / 100.00 \\
Rule-TVM & 0.33 / 61.67 & 0.00 / 46.67 & 0.67 / 100.00 & 0.67 / 99.33 \\
Semantic-TVM & 1.00 / \textbf{92.33} & 0.50 / \textbf{87.50} & 7.11 / 100.00 & 12.00 / 99.78 \\
\bottomrule
\end{tabular}
\end{table*}

\subsection{Experimental Setup}
\label{sec:experimental-setup}

\subsubsection*{Evaluation Pipelines}
\textbf{Memory-EHR.} This MIMIC-III query setting follows the EHRAgent-style code-generation workflow~\cite{shi2024ehragent}. The agent retrieves four examples by edit distance, generates a query program using the remote LLM, and executes it locally. Protected information includes patient identifiers, record bindings, and exact executable values present in historical examples but undisclosed by the current task.

\textbf{Memory-RAP.} This setting contains 500 WebShop memories and retrieves three examples by edit distance. In each trial, the agent constructs a search action from retrieved historical actions; a local search tool executes it and returns an observation. The current instruction and category are public, whereas an exact \texttt{search\_term} originating only from a historical action is protected. Memory-RAP comprises privacy-disclosure probes without independent ground-truth answers; Execution Success is its only utility metric.

\subsubsection*{Baselines and Methods}
\textbf{No Defense} sends raw retrieved memory; \textbf{Masking} applies one-way placeholder substitution without recovery or reprojection; \textbf{Rule-TVM} uses whole-field projection; \textbf{Classifier-TVM} uses the Semantic-TVM local model for whole-field selection and replacement; and \textbf{Semantic-TVM} uses it for span-level projection. Rule-TVM and Semantic-TVM share local resolution and reprojection.

\subsubsection*{Models, Runtime, and Metrics}
\textbf{Remote providers and runtime.} We use DeepSeek-V4-Flash and MiniMax-M2.5 at temperature 0 and a locally served predictor (\texttt{gemma4:e4b-it-qat}). Memory-EHR uses memory size 200 and $k=4$; Memory-RAP uses size 500 and $k=3$. Both use three seeds for the main comparisons; failure staging uses a single-seed DeepSeek EHR run.

\textbf{Evaluation metrics.} Leak Rate is the fraction of fully captured trials with a protected value in a remote request; incomplete trials are privacy-unknown and excluded. Execution Success and Task Success measure successful execution and final-answer correctness over all attempts.

\subsection{Results}
\label{sec:experimental-results}

Table~\ref{tab:main-results} summarizes the main comparison.

\subsubsection*{RQ1: Privacy and Executability}
No Defense exposes protected values in 78.50\% of EHR trials and all RAP trials (Table~\ref{tab:main-results}). Masking reduces EHR exposure but lacks recovery, yielding only 44.83\%--59.33\% Execution Success; on RAP, its unmediated return path leaves up to 22.67\% exposure. TVM combines low measured exposure with executable workflows: Leak Rate is at most 1.00\% on EHR and 12.00\% on RAP, while its span-level variant achieves 87.50\%--92.33\% EHR and at least 99.78\% RAP Execution Success.

\subsubsection*{RQ2: Projection Granularity}
\begin{figure}[!h]
\centering
\includegraphics[width=\columnwidth]{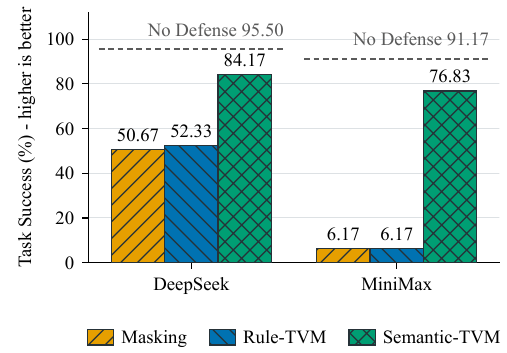}
\caption{Memory-EHR Task Success; dashed lines show No Defense.}
\label{fig:ehr-task-success}
\end{figure}

Span-level projection increases EHR Task Success from 52.33\% to 84.17\% on DeepSeek and from 6.17\% to 76.83\% on MiniMax (Fig.~\ref{fig:ehr-task-success}), while EHR Leak Rate remains at most 1.00\%. The gains occur across both remote backends despite their different whole-field baselines. Rule-TVM and Semantic-TVM share the resolver and reprojection path, allowing the projected representations to be compared under the same execution path.

Classifier-TVM changes field selection but retains whole-field replacement, yielding results near Rule-TVM (47.83\% versus 52.33\% on DeepSeek; 7.00\% versus 6.17\% on MiniMax). Failure staging further clarifies the gap: 78 of Rule-TVM's 94 unsuccessful DeepSeek trials end at formatting or execution, compared with 15 under span-level projection. Together, the cross-backend gains, classifier control, and shift in failure stages show that retained operators, relations, and argument structure help the remote model construct executable actions around protected references.

\subsubsection*{RQ3: Recovery and Local-Adapter Ablations}
The factorial ablation separates two functions of the TVM loop (Table~\ref{tab:ablation}). With local scene conditioning enabled, recovery increases EHR Task Success from 4.83\% to 84.17\% by connecting projected actions to exact values required by the executor. With recovery enabled, the local scene adapter reduces RAP Leak Rate from 89.11\% to 7.11\% by guiding span prediction across heterogeneous memory structures. Thus, recovery maps projected actions to exact local execution, while the adapter provides scene-specific cues to the local projector before serialization. The adapter itself remains local.

\begin{table}[!h]
\centering
\caption{Local adapter $\times$ Recovery ablation on DeepSeek over three seeds. Entries are EHR Task Success / RAP Leak Rate (\%).}
\label{tab:ablation}
\begin{tabular}{lcc}
\toprule
\textbf{Local adapter} & \textbf{Recovery off} & \textbf{Recovery on} \\
\midrule
Adapter off & 1.67 / 88.00 & 83.17 / 89.11 \\
Adapter on & 4.83 / 8.00 & 84.17 / 7.11 \\
\bottomrule
\end{tabular}
\end{table}

\FloatBarrier

\section{Related Work}
\label{sec:related}

Local mediation can reduce the private context disclosed to remote agents. SlotGuard~\cite{slotguard2026} rewrites provider-bound transcripts using typed slots and format-preserving substitutes, tracks cross-turn references, and restores bindings inside a trusted runtime. Its implementation and evaluation emphasize structural bindings and execution cues in repository-oriented workflows. PlanTwin~\cite{plantwin2026} instead exposes a constrained abstraction of a private environment for cloud-assisted planning.

MINIM~\cite{minim2026} predicts sensitivity and task-conditioned necessity for UI elements and applies Keep, Abstract, or Remove decisions before disclosure. Its controlled evaluation implements abstraction with standardized masking, while its framework permits richer implementations. Our study focuses on how to construct a recoverable span-level view of retrieved memories that retains surrounding operators, arguments, and relations for remote reasoning. We additionally evaluate the use of those references during local tool execution and the reprojection of subsequent observations.

\section{Limitations}
\label{sec:limitations}

Our exposure audit detects exact or normalized matches to manifest-listed values in fully captured provider-bound requests; it cannot establish semantic privacy or rule out inference from retained context. Trial-scoped handles reveal per-type cardinality and reference equality. TVM does not require a particular span detector, but our local predictor may miss or over-project content, especially in adversarial retrieved context. We evaluate two providers and two pipelines with different utility definitions; we do not measure local cost.

\section{Conclusion}
\label{sec:conclusion}

TVM separates remote-visible context from exact local execution state through projection, trusted recovery, and observation reprojection. Span-level projection preserves reasoning structure, while recovery connects the protected representation to executable utility with low measured exposure.

\clearpage
\section*{Compliance with Ethical Standards}

This study is a retrospective secondary analysis of the fully de-identified MIMIC-III database, accessed through PhysioNet under its credentialed data use agreement, together with a WebShop-derived memory set. No new human or animal data were collected and no human or animal subjects were involved. Because the MIMIC-III records are fully de-identified and were used under the PhysioNet data use agreement, no additional institutional review board approval or informed consent was required.

\section*{Acknowledgment}

This work was supported by Natural Science Foundation of Fujian Province (2024J08276, 2024J08278), Key Research Project for Young and Middle-aged Researchers by the Fujian Provincial Department of Education (JZ230044). The authors declare no competing financial or non-financial interests.

ChatGPT (OpenAI) assisted with the experimental and analysis code and with polishing the manuscript. The authors designed the method and the evaluation protocol, prepared all figures, ran all experiments, verified every reported result, and take full responsibility for this paper.

\bibliographystyle{IEEEbib}
\bibliography{tvm_refs}

\end{document}